\documentclass{article}

\usepackage{arxiv}

\usepackage[utf8]{inputenc} % allow utf-8 input
\usepackage[T1]{fontenc}    % use 8-bit T1 fonts
\usepackage{hyperref}       % hyperlinks
\usepackage{url}            % simple URL typesetting
\usepackage{booktabs}       % professional-quality tables
\usepackage{amsfonts}       % blackboard math symbols
\usepackage{nicefrac}       % compact symbols for 1/2, etc.
\usepackage{microtype}      % microtypography
\usepackage{lipsum}
\usepackage{graphicx}
\usepackage{amsmath}
\usepackage{makecell}
\usepackage{adjustbox}
\graphicspath{ {./images/} }

\title{AortaDiff: A Unified Multitask Diffusion Framework for Contrast-Free AAA Imaging}

\author{
Yuxuan Ou\textsuperscript{1}, Ning Bi\textsuperscript{2}, Jiazheng Pan\textsuperscript{3},
Jiancheng Yang\textsuperscript{4}\textsuperscript{5}, Boliang Yu\textsuperscript{2}, Usama Zidan\textsuperscript{2},\\
\textbf{Regent Lee\textsuperscript{2}, Vicente Grau\textsuperscript{1}} \\
\textsuperscript{1}Department of Engineering Science, University of Oxford, United Kingdom \\
\textsuperscript{2}Nuffield Department of Surgical Sciences, University of Oxford, United Kingdom \\
\textsuperscript{3}Technical University of Munich, Germany \\
\textsuperscript{4}ELLIS Institute Finland, Finland \\
\textsuperscript{5}Aalto University, Finland \\
\texttt{yuxuan.ou@trinity.ox.ac.uk}, \texttt{ning.bi@nds.ox.ac.uk}, \\
\texttt{jiazhen.pan@tum.de}, \texttt{jiancheng.yang@aalto.fi}, \texttt{boliang.yu@nds.ox.ac.uk}, \\
\texttt{usama.zidan@nds.ox.ac.uk}, \texttt{regent.lee@nds.ox.ac.uk},
\texttt{vicente.grau@eng.ox.ac.uk}
}

\begin{document}
\maketitle
\begin{abstract}
While contrast-enhanced CT (CECT) is standard for assessing abdominal aortic aneurysms (AAA) , the required iodinated contrast agents pose significant risks, including patient allergies and environmental harm. To reduce contrast agent use, methods have focused on generating synthetic CECT from non-contrast CT (NCCT) scans. However, most adopt a multi-stage pipeline that first generates images and then performs segmentation, which leads to error accumulation and fails to leverage shared information.
To address this, we propose a unified framework that generates synthetic CECT images from NCCT scans while simultaneously segmenting the aortic lumen and thrombus. Our approach integrates conditional diffusion models (CDM) with multi-task learning, enabling end-to-end joint optimization of image synthesis and anatomical segmentation. Unlike previous multitask diffusion models, our approach requires no initial predictions (e.g., a coarse segmentation mask), shares both encoder and decoder parameters across tasks, and employs a semi-supervised training strategy to learn from scans with missing segmentation labels, a common constraint in clinical data.
Evaluated on a cohort of 264 patients, our method consistently outperformed state-of-the-art single-task and multi-stage models. For image synthesis, it achieved a PSNR of 25.61 dB, compared to 23.80 dB from a single-task CDM. For segmentation, it improved the lumen Dice score to 0.89 from 0.87 and the challenging thrombus Dice score to 0.53 from 0.48 (nnU-Net). These segmentation enhancements led to more accurate clinical measurements, reducing the lumen diameter MAE to 4.19 mm from 5.78 mm and the thrombus area error to 33.85\% from 41.45\%. Code is at \url{https://github.com/yuxuanou623/AortaDiff}.
\end{abstract}

\begin{figure*}[t!]
    \centering
    \includegraphics[width=\textwidth]{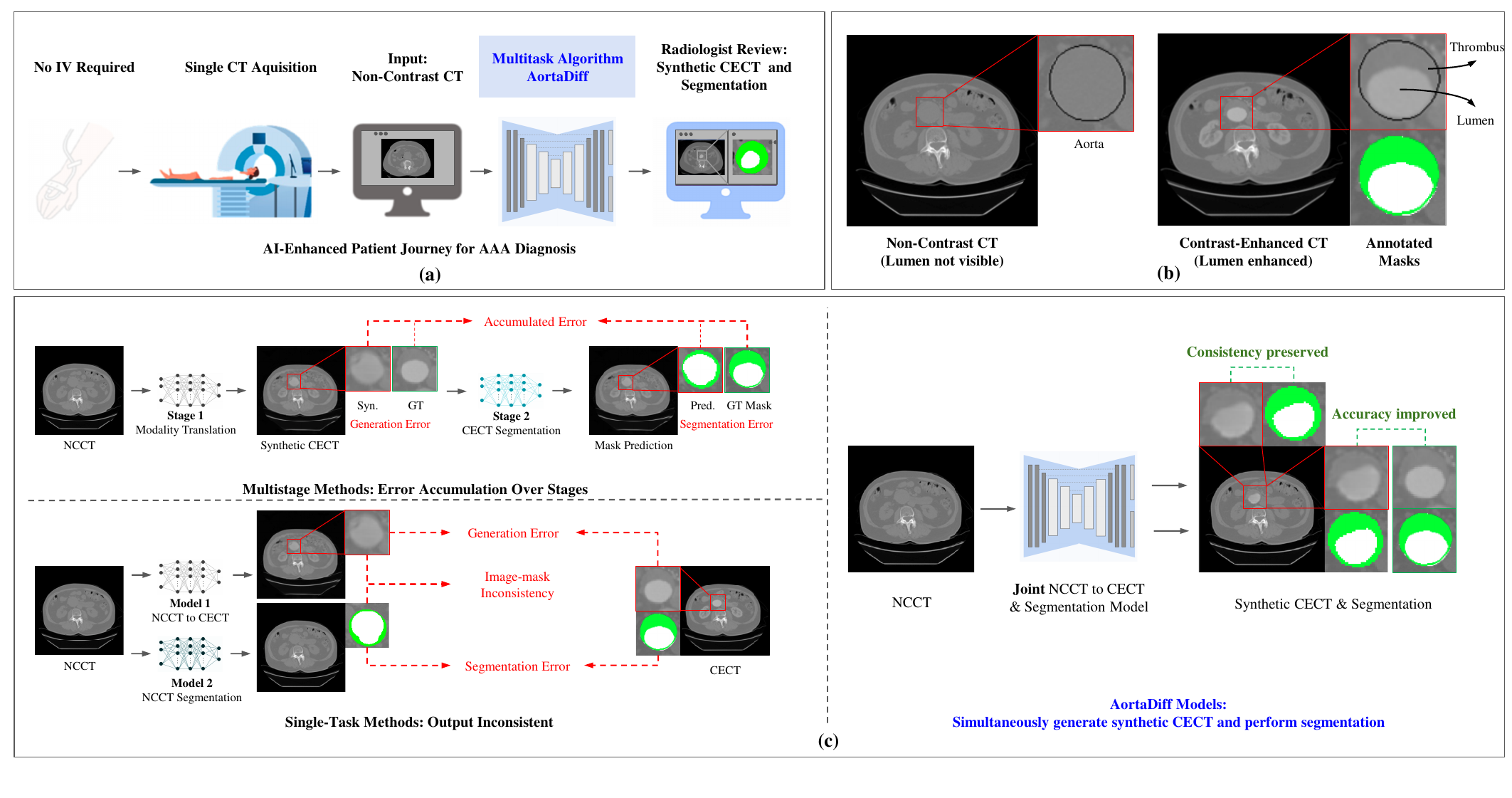}  % PDF spans both columns
    \caption{\textbf{Unified Multitask Framework for Contrast-Free AAA Assessment.} This figure illustrates the clinical motivation, CT imaging concepts, and advantages of our proposed multitask diffusion model (AortaDiff) over traditional approaches. \textbf{(a)} The proposed AI-enhanced clinical workflow enables diagnosis from a NCCT scan, eliminating the need for intravenous (IV) contrast agents. \textbf{(b)} A visual comparison highlights the diagnostic challenge: the aortic lumen is obscured in the NCCT but is clearly delineated in the CECT. \textbf{(c)} Our unified multitask model is contrasted with prior methods. Multistage pipelines risk accumulating synthesis errors into the final segmentation. Single-task approaches can produce segmentations that are inconsistent with the synthesized image. In contrast, our end-to-end model jointly generates a high-fidelity synthetic CECT and an accurate, consistent segmentation mask.}
    \label{fig1}
\end{figure*}

\section{Introduction}
% Motivated by these limitations, recent AI advancements aim to provide a safer, non-invasive alternative by digitally generating synthetic CECTs from readily available non-contrast CT (NCCT) scans \cite{santini2018synthetic, xie2021generation,chun2022synthetic, hu2022aorta,  chandrashekar2023deep, ristea2023cytran}.
% Therefore, an effective AI solution must perform two interrelated tasks: generate a realistic synthetic CECT to aid visual interpretation by radiologists and accurately segment the aortic lumen, allowing thrombus to be identified as the non-enhancing regions outside this segmented area.
 % To address this, some works introduce segmentation as an auxiliary task \cite{hu2022aorta, santini2018synthetic}, but the segmentation outputs are often threshold-dependent and not robust across varying patient anatomies and acquisition protocols.
Clinical assessment of vascular conditions like abdominal aortic aneurysms (AAA) traditionally relies on contrast-enhanced CT (CECT), where an intravenous iodinated contrast agent illuminates the aortic lumen to reveal abnormalities such as thrombus \cite{napoli2004abdominal, 10944207}. However, this procedure poses risks to patients with renal insufficiency or contrast allergies, involves uncomfortable needle insertions, and contributes to environmental iodine waste \cite{agentharm1, agentharm2, agentharm3, 10981129}. This has created a significant clinical need for an advanced method capable of performing AAA assessment solely on non-contrast CT (NCCT) scans. Such an approach is particularly challenging because thrombus and blood are nearly indistinguishable in NCCT, as shown in Figure~\ref{fig1} (b). Recent advances in deep learning offer a promising pathway to overcome this diagnostic challenge. An effective AI solution must therefore perform two interrelated tasks: (1) generate a realistic synthetic CECT to aid visual interpretation by radiologists, and (2) accurately segment the aortic lumen, allowing thrombus to be identified as the non-enhancing regions outside this segmented area. This dual-output approach provides radiologists with both a qualitative view of the anatomy from the synthetic image and immediate quantitative measurements of the abnormality from the segmentation mask, as illustrated in Figure~\ref{fig1} (a).

Recent AI advancements have primarily focused on generating synthetic CECTs from NCCT scans, exploring architectures from early CNNs \cite{santini2018synthetic} to GANs \cite{xie2021generation, chun2022synthetic, chandrashekar2023deep} and Transformers \cite{ristea2023cytran}. However, these works are constrained by a fundamental limitation: they frame the problem as a single-task, image-to-image translation. This overlooks the clinical reality that a visually plausible image is not the end goal, but an intermediate step towards quantitative analysis. Consequently, when segmentation is required, it is relegated to a separate, downstream model \cite{chun2022synthetic, chandrashekar2023deep}. This multi-stage pipeline is inherently flawed, as it is susceptible to error accumulation, where synthesis artifacts degrade segmentation accuracy, and prevents knowledge sharing between the two highly relevant tasks.

To address the shortcomings of multi-stage methods, some pioneering works have explored unified, multi-task frameworks. For instance, Hu et al. \cite{hu2022aorta} and UCAS \cite{zhu2025partial} proposed models that jointly perform CECT synthesis and segmentation. However, these early multi-task attempts are predominantly built upon GAN-based architectures. While innovative in their task formulation, they inherit the well-known challenges of GANs: they often produce overly smooth outputs that lack the sharp anatomical boundaries critical for clinical diagnosis and can suffer from training instability or mode collapse \cite{azarfar2023applications, 10.1007/978-3-032-06329-8_12}. Therefore, a significant gap remains for a framework that not only adopts a unified multi-task design but also leverages a more powerful generative model to ensure both high-fidelity synthesis and precise segmentation.

To address these limitations, we propose a unified multi-task diffusion framework, AortaDiff, that jointly performs image translation from NCCT to CECT and lumen segmentation. The core of our approach is a novel denoising U-Net architecture featuring a shared encoder-decoder backbone that ends in two different prediction heads: one for the denoised image and one for the segmentation mask. This single-backbone design allows the network to learn a rich, shared latent representation that includes both texture of contrast enhancement and precise anatomical information about lumen thrombus boundary. What's more, our model advances beyond prior multitask diffusion methods in two ways. First, it operates directly on the raw NCCT input, eliminating the need for any coarse initial predictions (e.g., a pre-computed coarse segmentation mask) during both training and inference. Second, to address the specific challenge of scarce lumen segmentation labels, we introduce a semi-supervised training strategy. This strategy incorporates a loss function that dynamically changes between supervised objectives on fully labeled data with an unsupervised reconstruction objective on image-only pairs. This allows our model to effectively leverage the entire dataset, even when only a subset of scans have lumen masks, enhancing its generalization capabilities under limited supervision and aligning with the practical constraints of clinical data acquisition.

We conducted a comprehensive evaluation of our framework on the Oxford Abdominal Aortic Aneurysm (OxAAA) dataset \cite{oxaaa}. Our unified model was benchmarked against state-of-the-art baselines and traditional multi-stage pipelines. The results demonstrate a consistent improvement across all tasks. For the challenging thrombus segmentation, our model increases the Dice score from 0.48 (nnU-Net) to 0.53, and for lumen segmentation, from 0.87 to 0.89. In the synthesis task, our framework achieves a PSNR of 25.61 dB, markedly outperforming the single-task CDM's 23.80 dB. Qualitatively, our jointly generated outputs exhibit sharper anatomical details and superior segmentation consistency. We also validate our method on two AAA clinical measurements and found that it yields more accurate morphological predictions for both lumen diameter and thrombus area, underscoring its potential for reliable clinical application.

In summary, the contributions of this work are threefold:
\begin{itemize}
    \item We propose a novel diffusion-based multi-task framework designed to move beyond image synthesis towards clinically actionable analysis. Based on the insight that a synthetic CECT is an intermediate step, our model jointly performs synthetic CECT generation and lumen segmentation. This approach enables a safer and more sustainable workflow for AAA imaging by reducing the need for contrast agent injection.
    
    \item To address the real-world challenge of missing lumen segmentation annotations, we introduce a semi-supervised strategy that enables effective learning from datasets where only a subset of image pairs contains lumen segmentation masks. This strategy improves both generation and segmentation performance beyond fully supervised training, leveraging cross-task consistency and multi-task representation learning.
    
    \item Experiments on the OxAAA dataset demonstrate that our model outperforms all single-task baselines and multi-stage pipelines on both generation and segmentation tasks. Our method is also more label-efficient than prior multi-task diffusion approaches, making it suitable for scalable deployment in resource-constrained healthcare environments.

\end{itemize}

\section{Related Work: Multi-task Diffusion Models}

Extending diffusion models for multi-task learning is an active area of research, with many foundational strategies first developed in general computer vision. For instance, CoDi \cite{codi} enables 'any-to-any' generation by aligning separate, pre-trained diffusers in a shared latent space via cross-attention. TaskDiffusion \cite{taskddpm} introduces a joint denoising process with a cross-task decoder to unify dense predictions, while DiffusionMTL \cite{diffmtl} uses a shared encoder that branches into task-specific decoders, conditioning the diffusion process on their initial outputs. Other works, like Shadow Diffusion \cite{shadowdiffusion}, require a coarse initial mask to guide the joint refinement process. Focusing on efficiency, MTU \cite{mtu} adapts pre-trained models by replacing network layers with smaller, task-specific feed-forward network  and proposing a routing mechanism. While these general-purpose frameworks are foundational, they are not tailored for the unique constraints of medical imaging, such as the need for high anatomical precision and handling of scarce clinical data.

Within the medical domain, a few works have begun to adapt these concepts. DiffAtlas \cite{diffatlas}, for example, applied a joint image-mask generation model to medical images but was limited by its requirement for full supervision. While all these works have significantly advanced multi-task diffusion, a critical gap remains for a framework that is simultaneously (1) architecturally unified and specialized for a medical task, (2) initialization-free and label-efficient via a semi-supervised scheme, and (3) validated on a high-stakes clinical application. Our work is designed to fill this specific void.

\section{Experiments}
We train and evaluate the proposed multi-tasking diffusion models, AortaDiff, on the Abdominal Aortic Aneurysm dataset released by the Oxford Abdominal Aortic Aneurysm (OxAAA) Study \cite{chandrashekar2023deep, oxaaa}. We assess two variants of our model: \textbf{AortaDiff-F}, trained with full supervision on fully-labeled data, and \textbf{AortaDiff-P}, trained with incomplete labels using our semi-supervised strategy on the entire dataset.

Performance is assessed on two fronts: synthetic CECT image generation and segmentation of the aortic lumen and thrombus. For each task, we compare our method against state-of-the-art single-task models trained with full supervision. In the segmentation task, we also evaluate the multi-stage baselines that first generate synthetic CECT and then segment the result. Both qualitative and quantitative results demonstrate the effectiveness of our proposed approach.

\subsection{OxAAA Dataset}

 The OxAAA dataset consists of 264 patient cases, each including spatially registered NCCT and CECT scans \cite{oxaaa}. Aorta segmentation labels are available for all cases, while only 65 cases additionally include expert-annotated lumen masks. We use 45 of the labeled cases for training and validation, and reserve 20 for testing. In our semi-supervised setting, the remaining 199 cases without lumen annotations are used during training.

\subsection{Dataset Preprocessing}
All CT scans used in our study are 3D volumes from which we extract 2D slices along the axial plane. To reduce redundancy from high inter-slice similarity, we retain one out of every three axial slices per volume, thereby balancing dataset size and information content. To ensure spatial alignment between NCCT and CECT scans, we retain only slice pairs with an aorta mask dice score above 0.9.

As lumen annotations are available only on the CECT scans, we transfer the corresponding lumen mask to the aligned NCCT slice. This process yields a total of 4,370 paired triplets, each consisting of an NCCT slice, its aligned CECT counterpart, and a corresponding lumen segmentation mask.

From these, we randomly select 2,823 triplets for training and 1,547 for testing. To avoid data leakage, all slices from the same patient are assigned exclusively to either the training or testing split. In addition, we include 7,366 NCCT-CECT slice pairs without lumen annotations, which are used during semi-supervised training.

Hounsfield Units are clipped to the range $[-1000, 1000]$, followed by min-max normalization to $[-1, 1]$ on a per-slice basis before input to the model.
 
\subsection{Experimental Settings}

We adopt the AdamW optimizer with a learning rate of \(1 \times 10^{-4}\) and no weight decay \cite{adamw}. The model is trained with a batch size of 16 and an input image matrix size of \(256 \times 256\). Training is conducted for a total of \(10^6\) iterations.

For the DDPM, we use a linear noise schedule with 1,000 diffusion timesteps.

\subsection{Evaluation Metrics}

For the image generation task, we evaluate the quality of the synthesized CECT slices using Peak Signal-to-Noise Ratio (PSNR), Structural Similarity Index Measure (SSIM) \cite{ssim}, and Learned Perceptual Image Patch Similarity (LPIPS) \cite{lpips}. Given the anatomical misalignment between the NCCT and CECT, particularly outside the aorta region, we focus our evaluation on the region of interest — the aorta. To reduce the influence of background artifacts and registration errors, we crop each slice to the smallest bounding box that encloses the aorta and compute the metrics within this cropped region.

\begin{figure*}[t]
    \centering
    \includegraphics[width=0.98\textwidth]{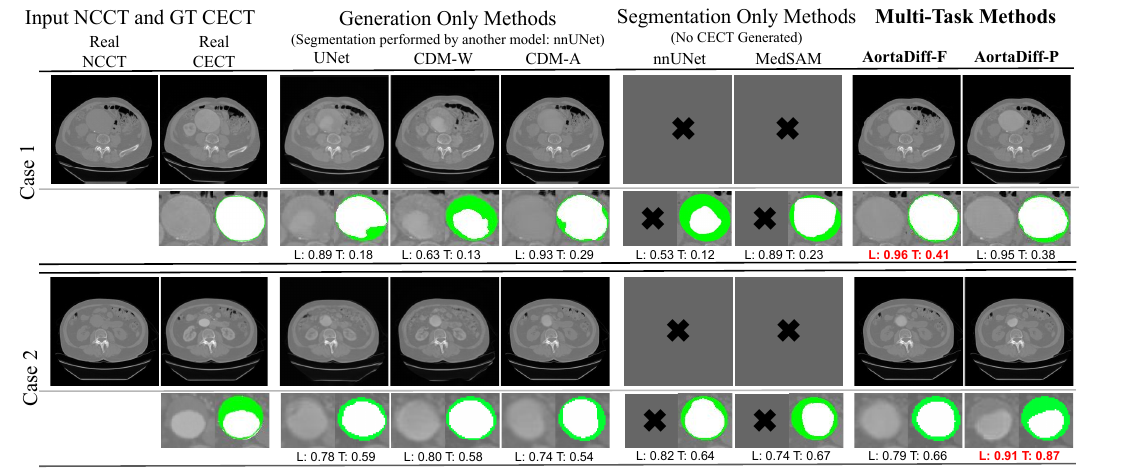} % Adjust scale as needed
    \caption{\textbf{Qualitative Comparison of Synthetic CECT Generation and Lumen and Thrombus Segmentation.} This figure illustrates two representative cases comparing generation-only (two-stage pipeline), segmentation-only, and our proposed multi-task diffusion methods. For each case, the top row displays the full generated image, while the bottom row provides a zoomed-in view of the aortic region with the corresponding segmentation result. In both cases, our multi-task methods achieve the most accurate segmentation with clearly defined boundaries and precise localization. Note that for generation-only methods, the segmentation is performed by nnU-Net on the generated synthetic CECT. For segmentation-only methods, no synthetic CECT is produced; thus, the corresponding CT images are shown in gray.}
    \label{fig3}
\end{figure*}

\subsection{Baselines}

\textbf{Synthetic CECT Generation.} To evaluate the performance of our method on synthetic CECT generation, we consider three baseline architectures: a U-Net \cite{unet}, a CycleGAN\cite{zhu2020unpairedimagetoimagetranslationusing} and a conditional diffusion model (CDM) \cite{adm}. The noise prediction network used in all diffusion models, including our proposed method, shares the same U-Net backbone for a fair architectural comparison. The CycleGAN experiments follow the default configurations provided in the official codebase.
For the CDM baseline, we evaluate two variants:

\noindent CDM-W (Whole Image):The entire NCCT slice is used as the conditioning input to generate the full CECT image.

\noindent CDM-A (Aorta-Restricted): Generation is restricted to the aortic region by conditioning only within the bounding box of the aorta.

This comparison allows us to assess whether focusing generation on the anatomically relevant region yields better results than generating the entire image. For the U-Net and CycleGAN baselines, the conditioning input is the whole NCCT image.

Due to misalignment between NCCT and CECT images, we replace the default MSE loss with a perceptual loss for all baseline models. In our ablation study, we demonstrate that incorporating perceptual loss improves performance across all models.

\noindent\textbf{Lumen Segmentation.} For the lumen segmentation task, we compare our method with nnU-Net \cite{nnunet}, a state-of-the-art medical image segmentation framework that achieves strong performance. Additionally, we include MedSAM \cite{medsam}, a pretrained vision foundation model for medical image segmentation. For MedSAM, we use the publicly released pretrained weights without any task-specific fine-tuning.

\subsection{Experiment Results}
\subsubsection{Quantitative Evaluation of CECT Synthesis}

Table~\ref{tab:comparison} presents quantitative results for CECT synthesis task, evaluated within the local aortic region. We observe that our proposed aorta-restricted generation strategy (CDM-A) outperforms both CDM-W, CycleGAN and U-Net across all metrics. CDM-W and CycleGAN underperform U-Net. This performance gap highlights the sensitivity of diffusion and GAN models to irrelevant regions in the conditioning input, such as those introduced by imperfect alignment between NCCT and CECT scans. These findings underscore the necessity of region-of-interest-focused generation to ensure stable performance in anatomically localized tasks.

Our proposed multi-task diffusion models (AortaDiff-F and AortaDiff-P) achieve the best and second-best performance across all evaluation metrics. This consistent improvement demonstrates that incorporating the segmentation task enhances the model’s ability to capture meaningful anatomical structure, thereby improving image synthesis quality. The results support the hypothesis that joint learning enables the model to leverage semantic guidance from the segmentation objective, leading to more accurate and anatomically coherent CECT generation.

\begin{table}[ht]
\centering
\begin{tabular}{l|c|c|c}
\toprule
\textbf{Method} & \textbf{PSNR (dB) $\uparrow$} & \textbf{SSIM $\uparrow$} & \textbf{LPIPS $\downarrow$} \\
\midrule
\multicolumn{4}{l}{\textit{Generation-Only Methods}} \\
UNet     & 24.40  & 0.7785 & 0.0815 \\
CycleGAN & 23.19  & 0.6535  & 0.1129 \\
CDM-W    & 23.80  & 0.7616 & 0.0859 \\
CDM-A    & 25.25  & 0.8137 & 0.0750 \\
\midrule
\multicolumn{4}{l}{\textit{Multi-Task Methods (Ours)}} \\
AortaDiff-F   & \textbf{25.61}  & \textbf{0.8385} & \underline{0.0671} \\
AortaDiff-P & \underline{25.48}  & \underline{0.8296} & \textbf{0.0606} \\
\bottomrule
\end{tabular}
\caption{\textbf{Quantitative Comparison on the NCCT-to-CECT Translation Task.} All metrics are computed within the local region surrounding the aorta. The best and second-best values are highlighted in \textbf{bold} and \underline{underlined}, respectively. Our multi-task diffusion models outperform baselines in all metrics.}
\label{tab:comparison}
\end{table}

\begin{table}[h]
\centering
\begin{tabular}{l|c|c}
\toprule
\textbf{Method} &
\makecell[c]{\textbf{Lumen}\\\textbf{Dice Score} $\uparrow$} &
\makecell[c]{\textbf{Thrombus}\\\textbf{Dice Score} $\uparrow$} \\
\midrule
\multicolumn{3}{l}{\textit{Segmentation-Only Methods}} \\
MedSAM     & 0.7479 & 0.4356 \\
nnUNet     & 0.8718 & 0.4750 \\
\midrule
\multicolumn{3}{l}{\textit{Generation-Only Methods (seg. by nnUNet)}} \\
UNet       & 0.8078 & 0.4604 \\
CycleGAN   &  0.6770     &0.4907        \\
CDM-W      & 0.7748 & 0.4339 \\
CDM-A      & \underline{0.8908} & 0.4914 \\
\midrule
\multicolumn{3}{l}{\textit{Multi-Task Methods (Ours)}} \\
AortaDiff-F     & 0.8887 & \underline{0.5182} \\
AortaDiff-P & \textbf{0.8933} & \textbf{0.5326} \\
\bottomrule
\end{tabular}
\caption{\textbf{Quantitative Comparison on the Lumen and Thrombus Segmentation Tasks.}  For generation-only methods, segmentation is performed by applying nnU-Net to the generated synthetic CECT images. Our semi-supervised multi-task diffusion models achieve superior performance on both tasks.}
\label{tab:seg}
\end{table}
\subsubsection{Quantitative Evaluation of Lumen and Thrombus Segmentation}

We evaluate lumen segmentation performance directly using each model’s predicted segmentation output. For thrombus segmentation, we first obtain aorta masks at test time using the pretrained nnU-Net described above. We then compute the thrombus mask by subtracting the predicted lumen region from the aorta region. This subtraction-based approach aligns with clinical definitions, where the thrombus is the area enclosed by the aortic wall but outside the contrast-filled lumen.

Table~\ref{tab:seg} presents Dice scores for both lumen and thrombus segmentation. We compare three groups of models: (1) segmentation-only baselines (MedSAM and nnU-Net), (2) generation-only methods, where nnU-Net is applied to synthetic CECT images, and (3) our proposed multi-task diffusion models (AortaDiff-F and AortaDiff-P).

The results reveal several key findings. First, for generation-only method, CDM-A outperforms both CDM-W, CycleGAN and U-Net in both tasks, confirming that ROI-constrained generation improves downstream segmentation quality. Second, AortaDiff-P achieves the highest Dice scores across both tasks, 0.8933 for lumen and 0.5326 for thrombus, exceeding even fully supervised segmentation-only methods.

Thrombus segmentation remains more challenging than lumen segmentation, as indicated by lower Dice scores. This is likely due to the thrombus often appearing as thin rings with fewer pixels. Nevertheless, the superior performance of our AortaDiff models suggests that jointly optimizing the generation and segmentation tasks allows the model to learn more accurate anatomical representations of the lumen and surrounding structures.

Importantly, our semi-supervised setup enables the model to benefit from training data without lumen segmentation labels. Even in such cases, the translation task facilitates learning structural cues that improve segmentation performance. This is particularly valuable in real-world clinical settings, where unlabeled imaging data are abundant, but high-quality annotations are expensive and labor-intensive to obtain. By learning jointly from both labeled and unlabeled data, our model demonstrates strong cross-task generalization.

\subsubsection{Qualitative Evaluation of CECT Synthesis and Lumen/Thrombus Segmentation}

Figure~\ref{fig3} presents two challenging clinical cases to qualitatively compare all baseline and proposed methods across both the CECT synthesis and segmentation tasks.

In Case 1, the reference CECT reveals a large, circular lumen with a thin but clearly visible mural thrombus encircling the vessel wall, corresponding to a circumferential thrombus morphology. UNet and CDM-W fail to synthesize the enhanced lumen correctly, exhibiting mottled textures and anatomically implausible aortic appearances. While CDM-A recovers a large circular lumen, it lacks contrast enhancement and fails to delineate the aortic boundary. In contrast, AortaDiff-F and AortaDiff-P produce sharper, anatomically realistic lumen enhancement and better preserve the aortic wall. On the segmentation task, AortaDiff models accurately localize the lumen, achieving the highest Dice scores among all methods. Although the thrombus in this case is relatively thin and thus difficult to delineate precisely, both AortaDiff variants outperform all comparison methods in identifying its presence and extent.

In Case 2, the thrombus extends along the posterolateral and lateral walls of the aorta, while the lumen appears as a bright, elliptical structure located in the inferior portion of the aortic cross-section. Only AortaDiff-P correctly synthesizes the shape, contrast, and position of the lumen, along with a clear interface between lumen and thrombus. For segmentation, other methods incorrectly predict the lumen as a large, centrally located circle, leading to under-segmentation of the thrombus. In contrast, AortaDiff models yield accurate segmentation of both the inferolaterally located lumen and the eccentric thrombus.

These results demonstrate that our multi-task diffusion models not only generate anatomically consistent synthetic CECT images but also deliver clinically meaningful segmentation results.

\subsection{Clinical Measurements for AAA}
Aneurysm morphology is important for predicting AAA progression \cite{oxaaa,Wanhainen2018-nk}. We evaluate two clinically relevant metrics, lumen diameter and thrombus area, on a random sample of ten test cases. As shown in Tables~\ref{tab:aorta-diameter-regression} and \ref{tab:thrombusarea}, our models produce accurate predictions on both metrics and outperform nnU-Net, indicating strong potential for clinical application.
\begin{table}[t]
\centering
\caption{\textbf{Lumen Diameter Regression on the Test Set.}  Best per column in \textbf{bold}.}
\label{tab:aorta-diameter-regression}
\setlength{\tabcolsep}{3pt}   % less side padding
\begin{tabular}{@{}lccc@{}} 
\toprule
\textbf{Method} & \textbf{MAE (mm) $\downarrow$} & \textbf{RMSE (mm) $\downarrow$} & \textbf{Pearson $r$ $\uparrow$} \\
\midrule
\multicolumn{4}{l}{\emph{Baseline}} \\
nnU-Net & 5.78 & 9.90 & 0.91 \\
\midrule
\multicolumn{4}{l}{\emph{Ours}} \\
AortaDiff-F & 4.70 & 7.05 & 0.93 \\
AortaDiff-P & \textbf{4.19} & \textbf{6.34} & \textbf{0.94} \\
\bottomrule
\end{tabular}
\end{table}

\begin{table}[t]
\centering
\caption{\textbf{Thrombus Area Prediction Error vs. Ground Truth.} Best per column in \textbf{bold}.}
\label{tab:thrombusarea}
\setlength{\tabcolsep}{3pt}
\begin{tabular}{@{}lc@{}}
\toprule
\textbf{Method} & \textbf{Mean $\pm$ SD (\%) $\downarrow$} \\
\midrule
\multicolumn{2}{@{}l}{\emph{Baseline}} \\
nnU\text{-}Net   & 41.45\% $\pm$ 29.53\% \\
\midrule
\multicolumn{2}{@{}l}{\emph{Ours}} \\
AortaDiff\text{-}F & 36.89\% $\pm$ 19.39\% \\
AortaDiff\text{-}P & \textbf{33.85\%} $\pm$ \textbf{16.52\%} \\
\bottomrule
\end{tabular}
\end{table}

\section{Ablation Study}
We conducted an ablation study to justify our design choices and provide insight into how these designs influence the results.

We evaluate the following design choices:
\begin{itemize}
  \item \textbf{Reconstruction Loss}: use MSE or perceptual loss for CECT image reconstruction. 
  \item \textbf{Mask representation}: use signed distance function to represent lumen mask instead of binary mask.
  \item \textbf{Multi-task loss combination}: use the uncertainty-aware multi-task loss instead of using the equal-weighted sum of segmentation and generation loss. 
\end{itemize}

From the reconstruction results in Table~\ref{tab:ablation0}, we observe that using LPIPS loss yields better reconstruction performance compared to MSE loss across all three models. This improvement is likely due to the background misalignments between our NCCT–CECT pairs, where a pixel-wise loss forces the model to learn these misalignments instead of focusing on perceptually meaningful features.
For both multi-task models, AortaDiff-F and AortaDiff-P, the results in Table~\ref{tab:ablation1} show that using an SDF-based mask representation yields better segmentation performance compared to a binary mask. In addition, Table~\ref{tab:ablation2} shows that using the uncertainty-aware loss leads to improved performance for both tasks compared to the equal-weight loss formulation.

\begin{table}[h!]
\centering
\small
\caption{Ablation results comparing different reconstruction losses, using either MSE or perceptual loss. }
\label{tab:ablation0}
\begin{adjustbox}{width=.60\linewidth}
\begin{tabular}{lccc}
\toprule
Reconstruction Loss & PSNR (dB) $\uparrow$ & SSIM $\uparrow$ & LPIPS $\downarrow$ \\
\midrule
CDM with MSE loss        & 20.87 & 0.7253 & 0.1077 \\
CDM with LPIPS loss      & 23.80 & 0.7616 & 0.0859 \\
CycleGAN with MSE loss     & 22.83 & 0.6492 & 0.1283 \\
CycleGAN with LPIPS loss   & 23.19 & 0.6535 & 0.1129 \\
U-Net with MSE loss        & 23.83 & 0.7602 & 0.0971 \\
U-Net with LPIPS loss      & 24.40 & 0.7785 & 0.0815 \\
\bottomrule
\end{tabular}%
\end{adjustbox}
\end{table}

\begin{table}[h!]
\centering
\small
\caption{Ablation results for different loss combination strategies.}
\label{tab:ablation1}
\begin{adjustbox}{width=.78\linewidth}
\begin{tabular}{l c c c c}
\toprule
Mask Representation & PSNR (dB) $\uparrow$ & SSIM $\uparrow$ & LPIPS $\downarrow$ & Lumen Dice $\uparrow$ \\
\midrule
AortaDiff-F Binary & 25.41 &0.8262  &0.0688  &0.8753  \\
AortaDiff-F SDF &25.61  &0.8385  &0.0671  &0.8887   \\
AortaDiff-P Binary & 25.47  &0.8198  &0.0661  &0.8760  \\
AortaDiff-P SDF &25.48   &0.8296  &0.0606  & 0.8933  \\
\bottomrule
\end{tabular}%
\end{adjustbox}
\end{table}

\begin{table}[h!]
\centering
\small
\caption{Ablation results for multi-task loss. All metrics are the same as in the main paper. PSNR, SSIM, and LPIPS are computed over the aorta region.}
\label{tab:ablation2}
\begin{adjustbox}{width=.98\linewidth}
\begin{tabular}{l c c c c}
\toprule
Multi-task Loss & PSNR (dB) $\uparrow$ & SSIM $\uparrow$ & LPIPS $\downarrow$ & Lumen Dice $\uparrow$ \\
\midrule
AortaDiff-F with Equal-Weight Loss &25.37  &0.8207  &0.0660  &0.8700  \\
AortaDiff-F with Uncertainty-Aware Loss &25.61  &0.8385  &0.0671  &0.8887  \\
AortaDiff-P with Equal-Weight Loss & 24.40  &0.7808  &0.0757  &0.8830  \\
AortaDiff-P with Uncertainty-Aware Loss &25.48   &0.8296  &0.0606  & 0.8933\\
\bottomrule
\end{tabular}%
\end{adjustbox}
\end{table}

\section{Conclusions}
In this work, we introduced AortaDiff, a unified multi-task diffusion framework for generating synthetic CECTs and segmenting the aortic lumen directly from NCCT scans. We demonstrated that having a diffusion model perform these tasks simultaneously enhances performance across both synthesis and segmentation. We also showed that an aorta-focused generation strategy with a background fusion technique can overcome significant data misalignment issues, while a direct, semi-supervised training scheme enables learning from annotation-scarce clinical datasets. Experiments show that our multitasking approach outperformed state-of-the-art single-task and multi-stage baselines on both synthesis and segmentation, leading to more accurate AAA clinical measurements. This research represents a meaningful step towards creating a ``digital contrast'' for CT. 

Despite these contributions, limitations remain. First, DDPM-based models have longer inference time than CycleGAN or U-Net based models, and the use of RePaint further increases inference time, posing challenges for clinical deployment. Second, our method currently requires lumen masks during training; since the model learns to generate the lumen, future research could investigate leveraging latent representations for unsupervised lumen prediction. Third, uncertainty estimation is crucial for clinical trust, motivating future work on quantifying uncertainty in both generation and segmentation.
\section*{Acknowledgements}
We acknowledge the contribution by the OxAAA Study Investigators (in particular Ashok Handa, Pierfrancesco Lapolla and Anirudh Chandrashekar) and the support by the Thames Valley Vascular Services, Oxfordshire, UK.

\bibliographystyle{unsrt}  
\bibliography{references}  %%% Remove comment to use the external .bib file (using bibtex).


\end{document}